\documentclass[11pt,letterpaper]{article}
\usepackage{labreport}
\usepackage{nicefrac}
\usepackage{adjustbox}
\usepackage{etoolbox}
\usepackage[section]{placeins}
\reportlab{}
\reporttype{Technical Report}
\reportnumber{}
\reportdate{}
\reportshorttitle{Physical-Field-Bridged Material Parameter Inference}
\reportcolors{183153}{167D8D}
\reportlinkcolor{2455A4}

\patchcmd{\maketitle}{\fontsize{24}{28}}{\fontsize{20}{24}}{}{}
\title{Physical-Field-Bridged Material Parameter\\Inference from Multi-View Deformation Videos}
\author{Linrui Shou\textsuperscript{1}\quad Jie Li\textsuperscript{2}\quad
  Yu Liu\textsuperscript{3}\quad Tao Huang\textsuperscript{1,*}}
\reportaffiliations{\textsuperscript{1}Shanghai Jiao Tong University\\
  \textsuperscript{2}University of Science and Technology Beijing\quad
  \textsuperscript{3}Peking University}
\reportauthornote{\textsuperscript{*}Corresponding author:
  \href{mailto:t.huang@sjtu.edu.cn}{t.huang@sjtu.edu.cn}.}
\reportresourcegroup{Project and Code}{%
  \reportresourceitem{Code repository}{https://github.com/3365538768/DeformFieldBench}{\nolinkurl{github.com/3365538768/DeformFieldBench}}}
\reportresourcegroup{Dataset}{%
  \reportresourceitem{DeformFieldBench}{https://huggingface.co/datasets/HandsomeHusky/DeformFieldBench}{\nolinkurl{huggingface.co/.../HandsomeHusky/DeformFieldBench}}}
\hypersetup{pdfauthor={Linrui Shou, Jie Li, Yu Liu, Tao Huang},
  pdfsubject={Physical-field-bridged material parameter inference}}
\AtBeginDocument{\hypersetup{pdfauthor={Linrui Shou, Jie Li, Yu Liu, Tao Huang}}}

\begin{document}
\maketitle
\begin{abstract}
Estimating simulation-ready material parameters from visual observations is crucial for deformable-object manipulation, yet remains challenging because material properties are only indirectly expressed through physical responses such as contact, deformation, recovery, and stress concentration. We introduce \emph{physical-field-bridged material inference}, a single-pass framework that treats dense physical responses as intermediate evidence between RGB deformation videos and a discrete constitutive type together with continuous material parameters. This design addresses a practical limitation of existing inverse-physics methods, which often rely on per-scene reconstruction, online tuning, or simulation-in-the-loop optimization, while feed-forward visual estimators can exploit appearance or category shortcuts. Given multi-view RGB videos, our model predicts force/contact masks, projected motion-flow fields, and projected stress-response maps, then conditions constitutive-type classification and parameter regression on these fields. To train and evaluate this formulation, we construct \textbf{DeformFieldBench}, a 3DGS--MPM learning environment with 5,000 simulations over 100 object shapes, five mechanically informative loading modes, and dense pixel-aligned physical supervision. Our method achieves the best median errors on all four continuous parameters, raises stress-hotspot recall from 0.0762 to 0.5121, and reaches 0.8676 SSIM/36.82 dB PSNR in same-simulator replay on unseen objects. Under a protocol-matched GSO evaluation, it performs inference and future-state prediction in approximately 30 seconds rather than the 15--120 minutes required by per-scene optimization; it also produces measurable replay agreement on five real-world sequences. These results support dense physical fields as an effective bridge for fast material inference, while cross-simulator and real-world tests expose remaining domain and identifiability gaps. The dataset is available on \href{https://huggingface.co/datasets/HandsomeHusky/DeformFieldBench}{Hugging Face}, and source code is available in the \href{https://github.com/3365538768/DeformFieldBench}{DeformFieldBench GitHub repository}.
\end{abstract}

\reportcontents

\section{Introduction}

Embodied agents need to predict how objects respond to physical interaction. For deformable objects, this requires estimating not only geometry or semantic category, but also material parameters such as Young's modulus, Poisson's ratio, density, and yield stress. These parameters are simulation-ready: once inferred, they can be inserted into a physics simulator to forecast deformation under new actions. Recent progress in inverse physics, neural rendering, and physics-aware 3D representations has started to bridge visual perception and physical simulation, enabling material estimation from videos or reconstructed geometry \cite{zhao2025masivmaterialagnosticidentificationvideos,caiGICGaussianInformedContinuum2024,liLearningPhysicsGrounded4D2026,chenEMPMEmbodiedMPM2026,jiangPhysTwinPhysicsInformedReconstruction2025}. Such a bridge is crucial for robotic manipulation, where perception should support actionable physical prediction rather than recognition alone.

However, current visual system-identification methods remain difficult to use as real-time perception modules. Many pipelines estimate material parameters through per-scene reconstruction \cite{li2023pac,chen2025vid2sim,caiGICGaussianInformedContinuum2024} or simulation-in-the-loop optimization \cite{zhao2025masivmaterialagnosticidentificationvideos,zhangPhysDreamerPhysicsBasedInteraction2024,chenEMPMEmbodiedMPM2026,jiangPhysTwinPhysicsInformedReconstruction2025}, repeatedly rolling out a simulator, comparing rendered future states with observations, and updating parameters. These methods are physically grounded, but are often computationally expensive, scene-specific, and dependent on reconstructed geometry, camera information, or iterative test-time tuning. A more scalable alternative is amortized feed-forward inference from RGB images or videos \cite{xuGaussianPropertyIntegratingPhysical2024,shuaiPUGSZeroshotPhysical2025,zhai2024physical,BMVC2016_39,48889}. Yet one-shot visual material inference is highly under-constrained: physical parameters are only indirectly expressed through appearance and motion, and existing feed-forward methods often rely on large vision-language models, semantic feature fields, interaction priors, or manually designed constraints. While useful, such priors can bias predictions toward object category, texture, or coarse material semantics. For deformable objects, a successful model should instead infer material properties from observed physical responses, such as contact, deformation propagation, recovery, and stress concentration.

\begin{figure}[tbp]
  \centering
  \includegraphics[width=1\linewidth]{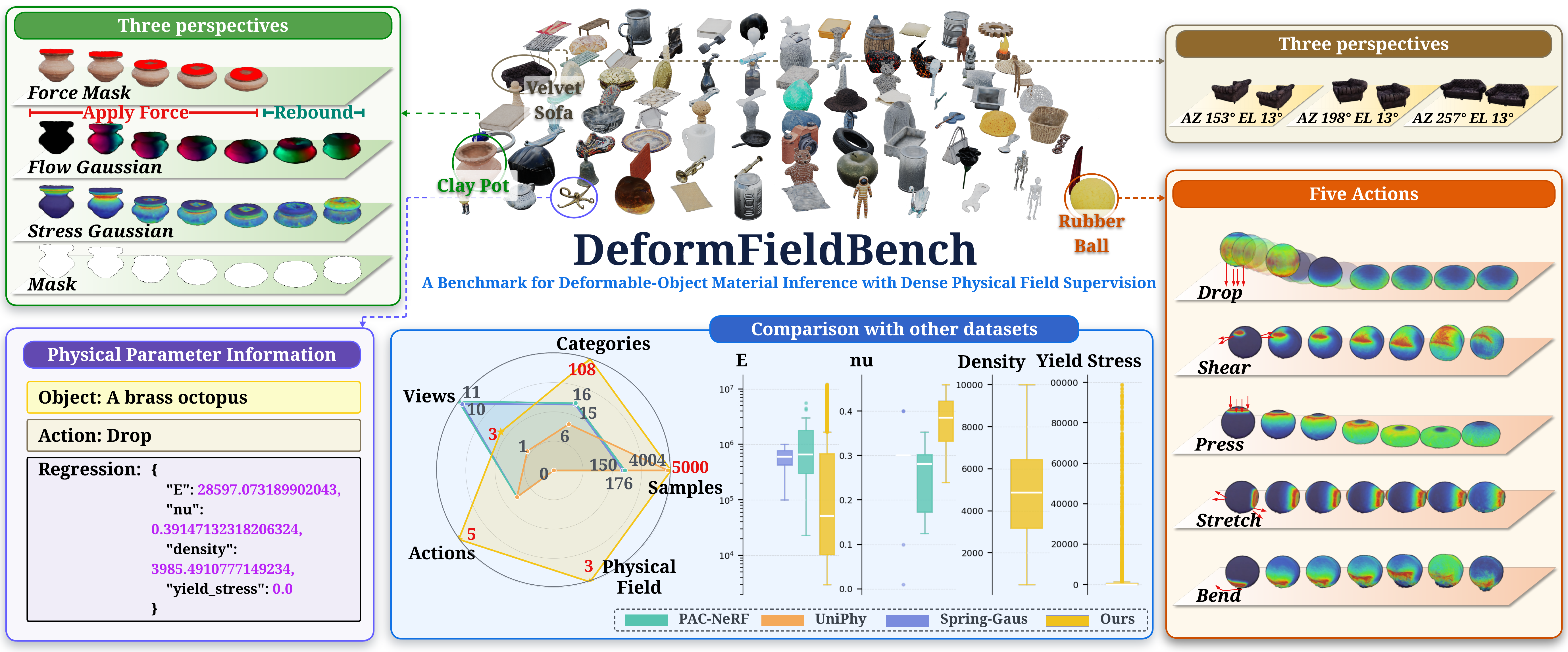}
  \caption{
  \textbf{Overview of DeformFieldBench.}
  Each sample combines three-view RGB observations with force/contact masks, projected flow, projected stress, object masks, a constitutive family, and continuous material parameters. The left panel illustrates pixel-aligned fields over loading and recovery, the center summarizes asset and parameter diversity relative to prior datasets, and the right shows the canonical views and five loading modes.}
  \label{fig:dataset}
\end{figure}

This motivates a missing ingredient: \textbf{dense physical supervision for learning the visual evidence behind material parameters}. Existing deformable-object datasets and physical-reasoning benchmarks \cite{bear2021physion,gan2020threedworld,lin2021softgym,bertiche2020cloth3d,antonova2021dynamic,zhang2024dofs,obristPokeFlexRealWorldDataset2025} often provide RGB videos, object states, or global labels, but rarely expose the intermediate physical fields that make material inference identifiable. Contact regions indicate where external forces are applied, motion fields reveal how deformation propagates, and stress responses highlight where force and deformation concentrate. Moreover, many benchmarks rely on limited geometries or action patterns dominated by dropping and impact, whereas compression, stretching, shearing, and bending reveal complementary properties such as stiffness, plasticity, and stress localization. Without diverse loading modes and dense physical fields, feed-forward models can easily collapse into black-box video-to-parameter regression.

To address this gap, we construct \textbf{DeformFieldBench}, a 3DGS--MPM simulation environment for learning simulation-ready material inference from multi-view RGB videos. It contains 5,000 simulations over 100 everyday object shapes, five mechanically informative loading modes, multiple material categories, and broad parameter ranges. Each sample provides synchronized three-view RGB deformation videos, object masks, force/contact masks, projected motion-flow fields, projected stress-response maps, and ground-truth material parameters. By combining the rendering efficiency of 3D Gaussian Splatting with the physical interpretability of MPM, DeformFieldBench provides pixel-aligned physical supervision while keeping the inference-time input purely visual. We use this benchmark not merely as a dataset contribution, but to study a central modeling question: how should dense physical fields be used to improve feed-forward material-parameter inference?

We introduce a physical-field-bridged multi-view model to answer this question. Unlike direct video regression, which maps RGB observations to material parameters through a single opaque representation, our model predicts intermediate physical responses and uses them as evidence for parameter estimation. Given multi-view RGB deformation videos, it extracts static geometric cues and dynamic deformation cues, fuses them across views, and decomposes the representation through a physics bottleneck. It then predicts force/contact regions, projected motion-flow fields, and projected stress-response fields. These predictions are not auxiliary outputs only: they are encoded and fused back into the material regressor, so that parameter estimation is conditioned on where the object is loaded, how motion propagates, and where stress concentrates. Since stress is less directly observable than contact and motion, we adopt a staged design that first learns visually grounded contact and flow prediction, then infers stress response, and finally regresses material parameters from the predicted fields and bottleneck representation.

The resulting framework provides an amortized feed-forward alternative to online system identification. It predicts a constitutive type, continuous material parameters, and dense physical-response fields in a single test-time forward pass, without per-scene reconstruction, camera-pose input, online tuning, or simulation-in-the-loop optimization. Experiments on object-level generalization show that our method achieves lower median errors on all four continuous parameters than VLM-based estimators and direct supervised video-regression baselines. It also substantially improves dense physical-field prediction, especially stress-hotspot localization. When the predicted constitutive model and parameters are fed back into the simulator, our method produces dynamically plausible future deformation on unseen objects. Protocol-matched, held-out-interaction, input-robustness, cross-domain, and real-world evaluations further characterize when this fast amortized inference transfers and where domain gaps remain.

In summary, our contributions are three-fold:
\begin{itemize}
    \item We construct \textbf{DeformFieldBench}, a 3DGS--MPM-based learning environment for feed-forward, simulation-ready material inference, with diverse objects, mechanically informative loading modes, multi-view RGB deformation videos, simulator-ready material labels, and dense physical-field supervision for force/contact, motion flow, and stress response.

    \item We propose a single-pass physical-field-bridged model that predicts dense contact, motion, and stress-response fields as intermediate evidence for constitutive-type classification and continuous material-parameter inference, avoiding per-scene reconstruction, camera-pose input, and simulation-in-the-loop optimization.

    \item We demonstrate improvements in median material-parameter accuracy, dense-field quality, object-level generalization, and same-simulator replay, and evaluate protocol-matched cross-benchmark transfer, unseen interactions, view and resolution sensitivity, lightweight domain adaptation, and quantitative real-world replay.
\end{itemize}

\section{Related Work}
\label{sec:related-work}

\paragraph{Physics-aware Gaussian representations and dynamic scene modeling}
Recent advances in 3D Gaussian representations have extended Gaussian primitives from static scene reconstruction to non-rigid dynamics and physically grounded modeling. Dynamic Gaussian manipulation methods study how Gaussian-based scene representations can support robotic interaction and non-rigid manipulation \cite{luManiGaussianDynamicGaussian2024}. Generative Gaussian methods further extend this representation to 4D asset generation and dynamic content synthesis \cite{renDreamGaussian4DGenerative4D2024, linDiffSplatRepurposingImage2025}. These methods provide flexible visual representations for dynamic objects, but they do not directly target the recovery of simulator-ready material parameters from deformation observations.

Another line of work couples Gaussian representations with physical simulation. PhysGaussian and related physics-integrated systems connect 3D Gaussians with MPM-based solvers, making it possible to generate or simulate physically plausible deformable dynamics from Gaussian primitives \cite{xiePhysGaussianPhysicsIntegrated3D2024, caoPhysX3DPhysicalGrounded3D2025}. These systems show that Gaussian representations can be made compatible with continuum simulation, which motivates our use of a 3DGS--MPM pipeline for dataset construction. However, their main focus is on forward simulation, reconstruction, or physically grounded generation, whereas our benchmark targets feed-forward material-parameter inference from multi-view RGB deformation videos.

Several recent methods also incorporate semantic or multimodal physical priors into Gaussian representations. Property-aware Gaussian methods use large multimodal models to infer coarse physical attributes or semantic-level material information \cite{xuGaussianPropertyIntegratingPhysical2024, shuaiPUGSZeroshotPhysical2025}. More recent physical Gaussian models and real-video-to-physics pipelines explore feed-forward 4D synthesis and physics-aware demonstration generation \cite{lvPhysGMLargePhysical2025, yangPhysWorldRealVideos2025}. These methods suggest that high-level priors are useful for physical reasoning, but they often struggle to capture fine temporal cues such as rebound, oscillation, and localized stress concentration. Our work instead uses dense stress, flow, and contact fields as explicit intermediate supervision for material inference.

\paragraph{Inverse physics and simulation-ready material inference}
System identification and inverse-physics methods aim to recover physical parameters from observed motion, deformation, or reconstructed geometry. Representative approaches include unified constitutive modeling for inverse simulation \cite{mittal2025uniphylearningunifiedconstitutive}, material-agnostic system identification from videos \cite{zhao2025masivmaterialagnosticidentificationvideos}, learning neural constitutive laws from motion observations \cite{maLearningNeuralConstitutive2023}, and Gaussian-informed continuum modeling for physical property identification and simulation \cite{caiGICGaussianInformedContinuum2024, liLearningPhysicsGrounded4D2026}. These methods are physically grounded and have demonstrated strong reconstruction or simulation performance in controlled settings.

Despite their strengths, many existing inverse-physics pipelines depend on assumptions that limit their use as real-time perception modules. Some methods require known geometry, depth observations, or reconstructed object states; others rely on iterative test-time optimization, where a simulator is repeatedly rolled out and matched to observations. This per-scene optimization can be expensive when the goal is to quickly infer physical parameters for downstream robotic manipulation. In addition, many methods optimize global parameters through future-state matching, but do not explicitly supervise intermediate physical responses. Without field-level supervision, a learned model may rely on object appearance, texture, or semantic category shortcuts rather than material-discriminative deformation cues. Our method addresses this issue by predicting motion flow, force/contact regions, and stress distributions before classifying the constitutive family and regressing Young's modulus, Poisson's ratio, density, and yield stress.

\paragraph{Feed-forward physics inference}
PIXIE predicts discrete material categories and continuous physical properties with supervised feed-forward learning from static multi-view appearance and geometry \cite{le2025pixie}. Our setting is complementary: it observes deformation videos under external interaction and explicitly predicts contact, motion-flow, and stress-response fields before material inference. This dynamic evidence is useful when objects with similar texture and geometry differ in stiffness, plasticity, density, or internal structure, properties that static appearance alone may leave ambiguous.

\paragraph{Datasets and benchmarks for physical scene understanding}
Datasets and simulation environments have played an important role in physical scene understanding. General-purpose environments such as Physion \cite{bear2021physion} and ThreeDWorld \cite{gan2020threedworld} provide standardized platforms for evaluating physical prediction and intuitive physical reasoning. However, these environments usually abstract away simulator-ready material parameters and internal physical response fields, which makes them less suitable for training models that infer numerical material properties.

Deformable-object benchmarks provide more relevant settings for manipulation and dynamics learning. SoftGym \cite{lin2021softgym}, Cloth3D \cite{bertiche2020cloth3d}, and DEDO \cite{antonova2021dynamic} focus on control-oriented deformable-object tasks, garment dynamics, or dynamic environments with deformable objects. These datasets are valuable for learning interaction policies and deformation dynamics, but they typically do not pair visual observations with explicit Young's modulus, Poisson's ratio, density, yield stress, or dense stress-field annotations.

Real-world resources provide complementary signals. DOFS \cite{zhang2024dofs} and PokeFlex \cite{obristPokeFlexRealWorldDataset2025} contain real deformable-object observations, geometry, or force-related measurements. These datasets are important for grounding deformable-object research in real measurements, but their scale, parameter coverage, and dense field annotations are limited for training generalizable material-inference models. Recent simulation-aligned pipelines such as Vid2Sim \cite{chen2025vid2sim} and PhysTwin \cite{jiangPhysTwinPhysicsInformedReconstruction2025} reconstruct geometry and physics from videos, but remain centered on per-scene reconstruction rather than large-scale benchmarking of feed-forward material inference.

DeformFieldBench is designed to complement these resources. It provides diverse object shapes, five mechanically informative loading modes, broad material-parameter ranges, and pixel-aligned physical fields for stress, flow, and force/contact regions. This enables systematic evaluation of whether models can infer simulation-ready material parameters from visual deformation evidence rather than from appearance alone.

\section{DeformFieldBench: A Learning Environment for Material Inference}

A central difficulty in feed-forward material-parameter inference is that the target variables are not directly observable from RGB appearance. Young's modulus, density, Poisson's ratio, and yield stress are expressed through how an object responds to external loading: where contact occurs, how deformation propagates, how the object recovers after release, and where stress concentrates. Existing datasets usually provide visual observations or reconstructed states, but rarely expose these intermediate physical responses in a form that can supervise learning. As a result, video-based models are often forced to learn a black-box mapping from appearance and motion to global material parameters, which makes them vulnerable to object-category or texture shortcuts.

We construct \textbf{DeformFieldBench} to study a different formulation: material inference as learning from dense physical responses. The benchmark is built on a 3DGS--MPM simulation pipeline and contains 5,000 simulations over 100 everyday object shapes, five loading modes (press, stretch, shear, bend, and drop), multiple material categories (jelly, metal, and plasticine), and broad material-parameter ranges. Each sample includes synchronized three-view RGB deformation videos, object masks, force/contact masks, projected motion-flow fields, projected stress-response maps, action labels, and ground-truth material parameters
\begin{equation}
    p = [E, \nu, \rho, \sigma_y],
\end{equation}
where $E$ is Young's modulus, $\nu$ is Poisson's ratio, $\rho$ is density, and $\sigma_y$ is yield stress. During inference, models are given only multi-view RGB videos; dense physical fields are used as intermediate learning targets. This setting allows us to evaluate whether a model can infer simulator-ready material parameters from visual deformation while being guided by physically meaningful evidence. Figure~\ref{fig:dataset} compares DeformFieldBench with related deformable-object and physical-reasoning datasets, highlighting its joint support for diverse object geometries, multiple mechanically informative loading modes, simulator-ready material labels, and dense projected physical-field supervision.

The current benchmark assumes one homogeneous constitutive description per object. Its dense fields provide spatial evidence for global system identification; they are not spatially varying material labels. The original videos are rendered at $512\times512$ resolution from three stationary canonical cameras distributed approximately horizontally around each normalized scene. Camera extrinsics are not provided to the model.

\subsection{Benchmark Design}

DeformFieldBench is built on a 3DGS--MPM simulation pipeline. 
Starting from a 3D Gaussian object representation, we convert visible Gaussian primitives into volumetric particles for MPM simulation, assign material parameters sampled from predefined ranges, and evolve the particles under external loading. 
The deformed particle states are then mapped back to Gaussian primitives and rendered into multi-view RGB videos and aligned physical-field videos. 
This design combines the rendering efficiency and appearance diversity of 3D Gaussian Splatting with the physical interpretability of MPM-based deformable simulation. A key design principle is to decouple visual appearance from material properties. By varying object geometry and surface style independently from $E$, $\nu$, $\rho$, and $\sigma_y$, the benchmark discourages models from predicting materials through category or texture shortcuts. 
Instead, models must rely on deformation behavior. 
To make such behavior informative, we include five loading modes: \textit{press}, \textit{stretch}, \textit{shear}, \textit{bend}, and \textit{drop}. 
These actions expose complementary physical cues, including local contact, compressibility, stiffness, tangential deformation, flexural recovery, inertia, and impact response. 
For actuator-driven actions, object-aware loading policies determine contact regions and trajectories from object geometry, and the simulation continues after release to capture rebound, oscillation, and residual deformation.

\subsection{Dense Physical Field Annotations}

The main distinction of DeformFieldBench is that RGB deformation videos are paired with dense, pixel-aligned physical fields. 
These fields are not privileged inputs at test time; they supervise the intermediate physical responses that should mediate material inference. 
\textbf{Force/contact masks} identify the regions affected by external actuation or contact, specifying where deformation is caused. 
\textbf{Projected motion-flow fields} are rendered from particle displacements between consecutive frames and describe how deformation propagates over the visible surface. 
\textbf{Projected stress-response maps} are computed from MPM stress and rendered into image space, revealing force concentration and yielding-related responses that may not be directly visible in RGB motion.

We export these annotations in image space rather than requiring full 3D particle trajectories or volumetric stress tensors. 
This keeps the learning problem aligned with visual perception, reduces storage and training cost, and still preserves the physical signals most relevant to material inference. 
The resulting supervision allows a model to learn not only what parameter values to predict, but also which physical evidence supports the prediction.

\paragraph{Efficient multi-modal field export}
A practical challenge in building dense physical supervision is that full 3D particle trajectories and per-particle stress tensors are expensive to store and difficult to use for video-based learning. 
We therefore export physical responses directly as image-space fields through Gaussian rendering. 
Specifically, particle motion is projected and encoded into rendered flow maps, while stress magnitude is mapped into rendered stress-response images. 
This design preserves dense, high-frequency physical signals while keeping the dataset lightweight and naturally aligned with RGB observations. 
It also makes DeformFieldBench scalable: compared with prior deformable physical-reasoning resources, our benchmark jointly supports broader object diversity, more loading types, wider parameter ranges, and stronger pixel-aligned physical supervision \cite{mittal2025uniphylearningunifiedconstitutive,obristPokeFlexRealWorldDataset2025,liLearningPhysicsGrounded4D2026,yangPhysWorldRealVideos2025}.

\subsection{3DGS--MPM Simulation Procedure}
\label{sec:simulation}

\subsubsection{Input Representation}

The input object is represented by a set of 3D Gaussian primitives
\begin{equation}
\mathcal{G}=\{(\mathbf{x}_p^0,\mathbf{\Sigma}_p^0,\alpha_p,\mathbf{c}_p)\}_{p=1}^{N},
\end{equation}
where $\mathbf{x}_p^0\in\mathbb{R}^3$ is the initial Gaussian center, $\mathbf{\Sigma}_p^0$ is its covariance, $\alpha_p$ is opacity, and $\mathbf{c}_p$ is the rendered color. Low-opacity Gaussians are discarded:
\begin{equation}
\alpha_p>\alpha_{\min}.
\end{equation}

The remaining Gaussian centers are normalized into the MPM simulation domain:
\begin{equation}
\mathbf{x}_p = \mathbf{1} + s\mathbf{R}(\mathbf{x}_p^0-\bar{\mathbf{x}}),
\end{equation}
where $\mathbf{R}$ is the preprocessing rotation, $s$ is a scale factor, $\bar{\mathbf{x}}$ is the object center, and $\mathbf{1}=(1,1,1)^T$ shifts the object near the center of the MPM box.

Each Gaussian is treated as an MPM particle with state
\begin{equation}
(\mathbf{x}_p,\mathbf{v}_p,\mathbf{F}_p,\mathbf{F}_p^{\ast},\mathbf{C}_p,V_p,m_p),
\end{equation}
where $\mathbf{v}_p$ is velocity, $\mathbf{F}_p$ is the elastic deformation gradient, $\mathbf{F}_p^{\ast}$ is the trial deformation gradient, $\mathbf{C}_p$ is the APIC affine velocity matrix, $V_p$ is particle volume, and $m_p$ is particle mass.

\subsubsection{Material Parameters}

The material parameters used in simulation are
\begin{equation}
E,\quad \nu,\quad \rho,\quad \sigma_y,\quad \mathbf{g},
\end{equation}
where $E$ is Young's modulus, $\nu$ is Poisson's ratio, $\rho$ is density, $\sigma_y$ is yield stress, and $\mathbf{g}$ is gravity.

The particle mass is
\begin{equation}
m_p=\rho V_p.
\end{equation}

The Lamé parameters are computed from $E$ and $\nu$:
\begin{equation}
\mu=\frac{E}{2(1+\nu)},
\qquad
\lambda=\frac{E\nu}{(1+\nu)(1-2\nu)}.
\end{equation}
Here $\mu$ controls shear resistance, while $\lambda$ controls volumetric resistance.

The material type determines the return mapping and stress law. In our setting, the main material behaviors are elastic jelly-like materials, metal-like von Mises plasticity, and plasticine-like plasticity with softening.

\subsubsection{Stress From Deformation}

At each substep, the solver first updates stress from the trial deformation gradient $\mathbf{F}_p^{\ast}$. A material-dependent projection gives
\begin{equation}
\mathbf{F}_p = \Pi(\mathbf{F}_p^{\ast}),
\end{equation}
where $\Pi$ is identity for elastic materials and a plastic return mapping for plastic materials.

For elastic materials, the solver uses a corotated stress model. Let
\begin{equation}
\mathbf{F}_p=\mathbf{U}_p\mathbf{S}_p\mathbf{V}_p^T,
\qquad
\mathbf{R}_p=\mathbf{U}_p\mathbf{V}_p^T,
\qquad
J_p=\det(\mathbf{F}_p).
\end{equation}
Here $\mathbf{U}_p,\mathbf{S}_p,\mathbf{V}_p$ come from singular value decomposition, $\mathbf{R}_p$ is the local rotation, and $J_p$ is the volume change. The Kirchhoff stress is
\begin{equation}
\boldsymbol{\tau}_p
=
2\mu(\mathbf{F}_p-\mathbf{R}_p)\mathbf{F}_p^T
+
\lambda J_p(J_p-1)\mathbf{I}.
\end{equation}

For plastic materials, the singular values of $\mathbf{F}_p^{\ast}$ are mapped to logarithmic strain
\begin{equation}
\boldsymbol{\epsilon}_p=\log(\boldsymbol{s}_p),
\end{equation}
where $\boldsymbol{s}_p$ contains the singular values. The deviatoric strain is
\begin{equation}
\hat{\boldsymbol{\epsilon}}_p
=
\boldsymbol{\epsilon}_p
-
\frac{1}{3}\operatorname{tr}(\boldsymbol{\epsilon}_p)\mathbf{1}.
\end{equation}
Plastic yielding occurs when the deviatoric part exceeds the yield stress. A simple return step is
\begin{equation}
\Delta\gamma
=
\|\hat{\boldsymbol{\epsilon}}_p\|
-
\frac{\sigma_y}{2\mu},
\end{equation}
\begin{equation}
\boldsymbol{\epsilon}_p
\leftarrow
\boldsymbol{\epsilon}_p
-
\frac{\Delta\gamma}{\|\hat{\boldsymbol{\epsilon}}_p\|}
\hat{\boldsymbol{\epsilon}}_p.
\end{equation}
Thus, $E$ and $\nu$ determine stiffness through $\mu,\lambda$, while $\sigma_y$ controls when plastic deformation starts.

\subsubsection{One MPM Substep}

Let $\Delta x$ be the grid spacing and $\Delta t$ be the substep size. Each particle transfers mass and momentum to nearby grid nodes. Let $w_{ip}$ be the interpolation weight between particle $p$ and grid node $i$, and $\nabla w_{ip}$ its spatial gradient.

\textbf{Particle to grid.}
\begin{equation}
m_i=\sum_p w_{ip}m_p,
\end{equation}
\begin{equation}
\mathbf{p}_i
=
\sum_p
w_{ip}m_p
\left(
\mathbf{v}_p+\mathbf{C}_p(\mathbf{x}_i-\mathbf{x}_p)
\right)
-
\Delta t
\sum_p
V_p\boldsymbol{\tau}_p\nabla w_{ip}.
\end{equation}
The first term transfers particle momentum to the grid. The second term applies elastic internal force from stress.

\textbf{Grid update.}
\begin{equation}
\mathbf{u}_i=\frac{\mathbf{p}_i}{m_i}+\Delta t\,\mathbf{g},
\end{equation}
where $\mathbf{u}_i$ is the grid velocity.

\textbf{Contact and actuation.}
For a moving cuboid actuator with center $\mathbf{q}(t)$, half-size $\mathbf{a}$, and velocity $\mathbf{u}_b$, grid nodes inside the cuboid satisfy
\begin{equation}
|\mathbf{x}_i-\mathbf{q}(t)|<\mathbf{a}.
\end{equation}
Their velocity is overwritten by
\begin{equation}
\mathbf{u}_i \leftarrow \mathbf{u}_b.
\end{equation}
The actuator moves as
\begin{equation}
\mathbf{q}(t+\Delta t)=\mathbf{q}(t)+\Delta t\,\mathbf{u}_b.
\end{equation}
This is how press, shear, stretch, and bend actions impose external motion. For drop, gravity and ground contact dominate the motion.

\textbf{Grid to particle.}
The updated grid velocities are interpolated back:
\begin{equation}
\mathbf{v}_p^{n+1}=\sum_i w_{ip}\mathbf{u}_i,
\end{equation}
\begin{equation}
\nabla\mathbf{v}_p=\sum_i \mathbf{u}_i(\nabla w_{ip})^T,
\end{equation}
\begin{equation}
\mathbf{F}_p^{\ast,n+1}
=
(\mathbf{I}+\Delta t\nabla\mathbf{v}_p)\mathbf{F}_p^n,
\end{equation}
\begin{equation}
\mathbf{x}_p^{n+1}
=
\mathbf{x}_p^n+\Delta t\,\mathbf{v}_p^{n+1}.
\end{equation}

\subsubsection{Stress Output}

The solver stores Kirchhoff stress $\boldsymbol{\tau}_p$. For output, it is converted to Cauchy stress:
\begin{equation}
\boldsymbol{\sigma}_p
=
\frac{1}{J_p}\boldsymbol{\tau}_p,
\qquad
J_p=\det(\mathbf{F}_p^{\ast}).
\end{equation}

The scalar stress visualization uses von Mises stress:
\begin{equation}
\sigma_{\mathrm{vm}}
=
\sqrt{
\frac{1}{2}
\left[
(\sigma_{11}-\sigma_{22})^2
+
(\sigma_{22}-\sigma_{33})^2
+
(\sigma_{33}-\sigma_{11})^2
+
6(\sigma_{12}^2+\sigma_{23}^2+\sigma_{13}^2)
\right]
}.
\end{equation}
This scalar is normalized to $t_p\in[0,1]$ and used to color the Gaussian:
\begin{equation}
\mathbf{c}_p^{\mathrm{stress}}
=
\beta\mathbf{c}_p+(1-\beta)\operatorname{ColorMap}(t_p).
\end{equation}

\subsubsection{Flow Output}

For each rendered view, the current particle position is projected to image coordinates:
\begin{equation}
\mathbf{u}_p(t)=\Pi(\mathbf{x}_p(t))=(u_p(t),v_p(t)).
\end{equation}
The optical-flow-like signal is the screen displacement between two output frames:
\begin{equation}
\Delta\mathbf{u}_p(t)
=
\mathbf{u}_p(t)-\mathbf{u}_p(t-\Delta t_{\mathrm{out}}).
\end{equation}
Its direction and magnitude are
\begin{equation}
\theta_p=\operatorname{atan2}(\Delta v_p,\Delta u_p),
\qquad
r_p=\|\Delta\mathbf{u}_p\|_2.
\end{equation}
The direction $\theta_p$ is encoded as hue, and the magnitude $r_p$ is encoded as brightness/opacity before Gaussian rasterization.

\subsubsection{Force and Contact Mask Output}

The force mask is defined from the active moving cuboid actuators. A particle is marked as being in the force/contact region if
\begin{equation}
|\mathbf{x}_p(t)-\mathbf{q}(t)|<\mathbf{a}.
\end{equation}
Thus the binary force mask is
\begin{equation}
M_p^{\mathrm{force}}(t)
=[|\mathbf{x}_p(t)-\mathbf{q}(t)|<\mathbf{a}].
\end{equation}
For single-channel mask rendering, only selected particles are visible:
\begin{equation}
\alpha_p^{\mathrm{force}}=M_p^{\mathrm{force}}\alpha_p,
\qquad
\mathbf{c}_p^{\mathrm{force}}=(1,1,1).
\end{equation}

The object mask uses all visible object Gaussians:
\begin{equation}
\alpha_p^{\mathrm{object}}=\alpha_p,
\qquad
\mathbf{c}_p^{\mathrm{object}}=(1,1,1).
\end{equation}

\subsubsection{Rendering Back to 3DGS}

After each selected simulation frame, the MPM particle positions are transformed back to the original 3DGS coordinate system:
\begin{equation}
\mathbf{x}_p^{\mathrm{world}}(t)
=
\bar{\mathbf{x}}
+
s^{-1}\mathbf{R}^{-1}
(\mathbf{x}_p(t)-\mathbf{1}).
\end{equation}
The Gaussian covariance can be updated by deformation:
\begin{equation}
\mathbf{\Sigma}_p(t)
=
\mathbf{F}_p(t)\mathbf{\Sigma}_p^0\mathbf{F}_p(t)^T.
\end{equation}
The updated Gaussian centers, covariances, colors, and opacities are then rendered to produce RGB frames, stress frames, flow frames, force masks, and object masks.

\subsection{Tasks and Evaluation Protocol}

The primary task is to infer a constitutive type $\hat{c}$ and continuous material parameters from synchronized three-view RGB videos: $\hat{p} = [\hat{E}, \hat{\nu}, \hat{\rho}, \hat{\sigma}_y]$.
We evaluate $E$, $\rho$, and $\sigma_y$ in log space due to their large dynamic ranges, and evaluate $\nu$ in raw space. Besides per-parameter errors, we report sample-wise parameter error to measure whether a method predicts a coherent parameter set for each simulation sample.

We use an object-level split, with 90 objects for training and 10 unseen objects for testing, to evaluate generalization beyond memorized shapes or appearances. We also evaluate dense field prediction: contact-region quality for force masks, coverage and velocity-aware quality for motion flow, and hotspot localization for stress response. Finally, we perform simulation replay by feeding the predicted constitutive type and parameters back into the simulator and comparing the rendered deformation with the ground truth. This replay protocol is important because different parameter combinations may induce similar visible motion; a useful estimate should therefore be not only numerically accurate, but also dynamically plausible. Together, parameter accuracy, field quality, object-level generalization, and replay performance provide the basis for evaluating the physical-field-bridged model introduced next. Details are reported in Section \ref{sec:evaluation-protocol}.

\subsection{Dataset Release and Usage Documentation}
\label{sec:dataset-release}

\paragraph{Dataset name}
DeformFieldBench: Physical-Field Material Parameter Dataset with 5,000 3DGS--MPM simulations.

\paragraph{Hosting and access}
The dataset artifacts are available on \href{https://huggingface.co/datasets/HandsomeHusky/DeformFieldBench}{Hugging Face}. Source code, documentation, and release updates are available in the \href{https://github.com/3365538768/DeformFieldBench}{DeformFieldBench GitHub repository}. The released files include rendered multi-view RGB deformation videos, projected physical-response fields, masks, action labels, material parameters, and train/test split metadata.

\paragraph{Dataset purpose}
DeformFieldBench is designed for simulation-ready material-parameter inference of deformable objects from multi-view RGB deformation videos. It supports both global parameter regression and dense physical-response prediction.

\paragraph{Dataset scale}
The dataset contains 5,000 simulation samples over 100 object shapes. Each object is simulated under five loading modes: press, stretch, shear, bend, and drop. Each sample contains $512\times512$ observations from three stationary canonical cameras distributed approximately horizontally around the normalized scene, together with aligned physical-field annotations.

\paragraph{Modalities and annotations}
Each sample contains:
(1) multi-view RGB deformation videos;
(2) object masks;
(3) force/contact masks;
(4) projected motion-flow fields;
(5) projected stress-response maps;
(6) action labels;
(7) a ground-truth constitutive family; and
(8) continuous material parameters, including Young's modulus, Poisson's ratio, density, and yield stress.

\paragraph{Generation pipeline}
The dataset is generated using a 3DGS--MPM simulation pipeline. Starting from 3D Gaussian object representations, we densify the object volume into MPM particles, assign material parameters, apply controlled mechanical loading, and render RGB videos and image-space physical-response fields through the Gaussian rendering pipeline.

\paragraph{Splits}
We use an object-level split to evaluate generalization. Specifically, 90 objects are used for training and 10 completely unseen objects are reserved for testing. This prevents identical object shapes from appearing in both training and test sets.

\paragraph{Evaluation tasks}
The benchmark supports:
(1) constitutive-type classification and material-parameter regression;
(2) dense stress-response prediction;
(3) projected motion-flow prediction;
(4) force/contact mask prediction;
(5) simulation replay evaluation using the predicted constitutive type and material parameters.

\paragraph{Known limitations}
The dataset is synthetic and generated by one 3DGS--MPM pipeline. It assumes a single homogeneous material description per object and uses a canonical three-camera layout. The stress and flow annotations are image-space projections rather than full volumetric fields, and normalized stress maps do not provide absolute stress values. High-stiffness metallic materials may therefore be difficult to distinguish when constant-velocity loading induces saturated permanent deformation. Cross-simulator results further show sensitivity to damping, filling, integration, contact, and boundary-condition implementations. Future versions could include heterogeneous materials, randomized camera and simulator configurations, calibrated external force, per-particle absolute stress, and tactile or force-torque signals.

\paragraph{License and usage}
Users should refer to the \href{https://huggingface.co/datasets/HandsomeHusky/DeformFieldBench}{Hugging Face dataset page} for the license, terms of use, and download instructions.

\section{Physical-Field-Bridged Material Inference}
\label{sec:method}

\begin{figure}[tbp]
  \centering
  \includegraphics[width=1\linewidth]{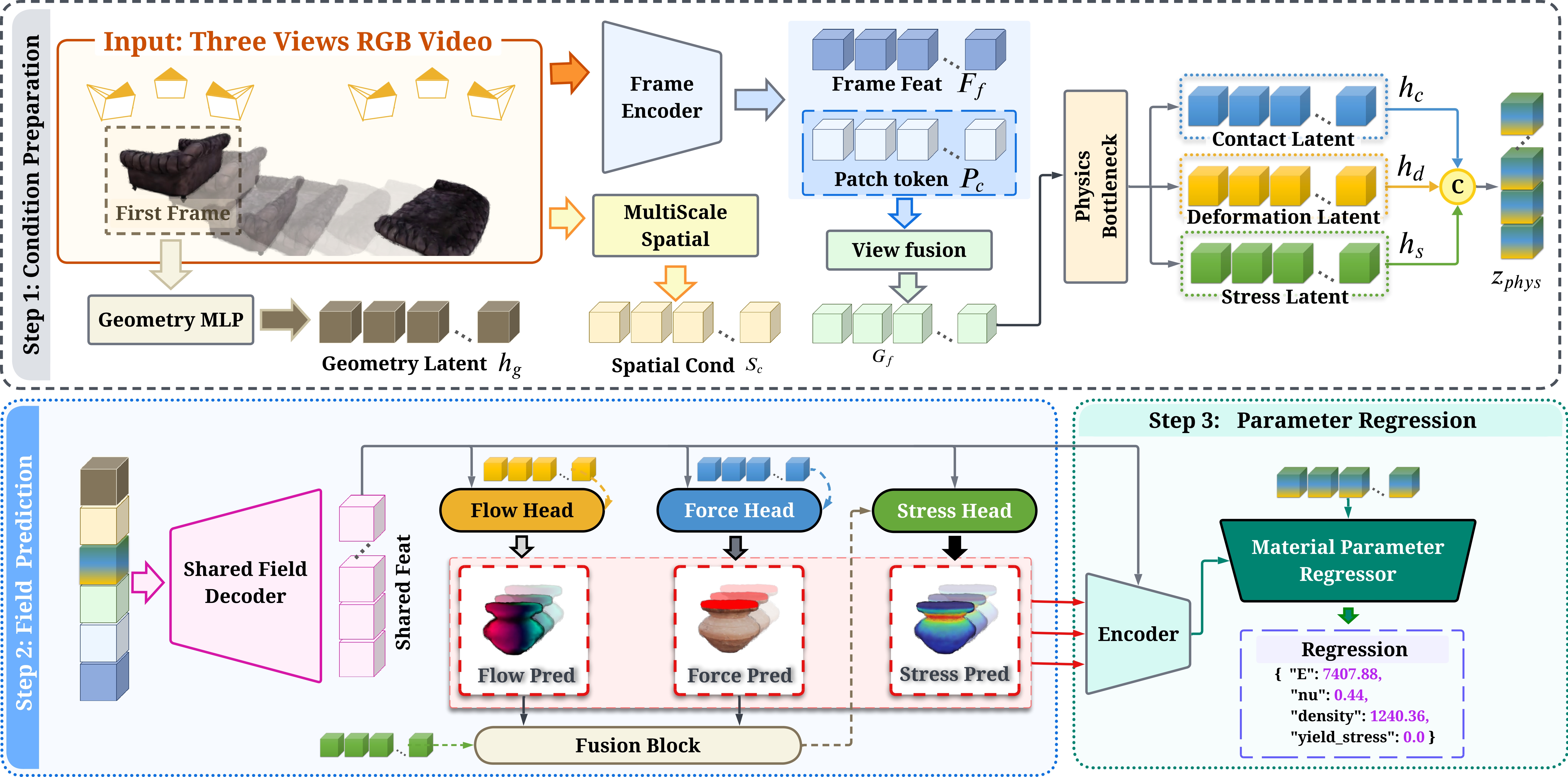}
  \caption{
  \textbf{Overview of the proposed model.}
  Multi-view RGB deformation videos are encoded into static and dynamic features, decomposed through a physics bottleneck, and decoded into dense motion-flow, force/contact, and stress fields. The predicted fields are encoded and fused with the bottleneck for material inference. The implemented constitutive-type classifier runs in parallel with the illustrated continuous-parameter regressor and is omitted from the diagram for clarity.}
  \label{fig:model}
\end{figure}

Our feed-forward model maps multi-view RGB deformation videos to simulation-ready material parameters through explicit physical evidence. Direct regression can entangle material cues with category and texture, while fields trained only as auxiliary outputs need not influence the final prediction. As shown in Fig.~\ref{fig:model}, our model instead fuses frame-, patch-, and multi-scale features, factorizes them through a physics bottleneck for contact, deformation, and stress, and supervises the corresponding dense fields. Encodings of the predicted fields are then fused back into the bottleneck, conditioning constitutive and parameter inference on where loading occurs, how motion propagates, and where stress concentrates.

Given synchronized RGB videos from $V$ views, $X = \{ I_{v,t} \mid v=1,\ldots,V,\; t=1,\ldots,T \}$,
our model predicts dense contact, motion, and stress fields,
$\hat{\mathcal{Y}} = \{\hat{M}, \hat{U}, \hat{S}\}$,
a discrete constitutive type $\hat{c}$, and global material parameters $\hat{p} = [\hat{E}, \hat{\nu}, \hat{\rho}, \hat{\sigma}_y]$.
Here $\hat{M}$ denotes the force/contact mask, $\hat{U}$ denotes projected motion flow, and $\hat{S}$ denotes the projected stress-response map. 
Only RGB videos are required at inference time; dense fields supervise the internal representation during training.

\subsection{Multi-view Visual Encoding}

Material inference requires static geometry, which affects load distribution, and dynamic cues such as deformation, rebound, oscillation, and residual displacement. A pretrained DINOv2 backbone extracts frame features and patch tokens; a temporal adapter aggregates dynamics within each view, and a cross-view transformer produces a global multi-view condition. Patch tokens and multi-scale RGB features remain as local conditions for dense decoding, preserving details that global features alone would discard.

For each frame in each camera view, we use a pretrained DINOv2 backbone to extract frame-level visual features. Let \(x_{v,t}\) denote the RGB frame at view \(v\) and time \(t\). The frame encoder produces
\begin{equation}
f_{v,t} = \Phi_{\mathrm{enc}}(x_{v,t}),
\end{equation}
where \(\Phi_{\mathrm{enc}}\) denotes the frozen visual backbone. In addition to the global frame token, patch-level tokens are also retained and later used as local spatial conditions for dense field prediction.

To model temporal deformation within each view, the frame features are passed through a lightweight Transformer temporal adapter with learnable temporal embeddings:
\begin{equation}
g_v = \mathrm{Pool}_t\left(\mathrm{Trans}_t(f_{v,1:T} + e_{1:T})\right),
\end{equation}
where \(g_v\) is the view-level dynamic representation. The resulting view features are further fused by a cross-view Transformer. Before fusion, a learnable view embedding is added to each view representation:
\begin{equation}
z = \Psi_{\mathrm{view}}\left(\{g_v + a_v\}_{v=1}^{V}\right).
\end{equation}
Here, \(z\) is the fused multi-view representation, which serves as the global visual condition for subsequent physical reasoning. Meanwhile, multi-scale RGB features and patch-level DINO features are extracted to provide local geometry and deformation cues for dense field decoding.

\subsection{Physics Bottleneck}

Instead of sending the fused video representation directly to all output heads, we introduce a physics bottleneck that decomposes it into contact-, deformation-, and stress-related latent factors:
\begin{equation}
    h_c, h_d, h_s = \Phi_{\mathrm{phys}}(z), 
    \qquad 
    z_{\mathrm{phys}} = [h_c; h_d; h_s],
\end{equation}
where $z$ is the fused multi-view visual feature. 
This bottleneck reflects the structure of the inference problem. 
Contact describes where external loading enters the object, deformation describes how the object moves under that loading, and stress describes the internal response that links visible motion to material parameters.

The decomposition organizes visual evidence for field-specific heads instead of exposing all outputs to one opaque representation: contact marks where loading enters, deformation describes the resulting motion, and stress connects that response to material behavior.

\subsection{Dense Physical Field Prediction}

A shared spatiotemporal decoder combines the global condition, physical latents, frame and patch features, and multi-scale spatial features; task-specific heads predict force/contact, motion flow, and stress response. The heads follow an observability hierarchy: contact and flow are grounded in actuator location and visible deformation, whereas stress is less directly observable and is therefore conditioned on both predicted responses. This sequential dependency ties stress to loading and deformation propagation rather than surface appearance alone. Object masks suppress background responses, especially for sparse stress and flow signals.

\subsubsection{Dense Physical Field Decoder}

Dense physical fields are predicted by a shared conditional spatiotemporal decoder. The decoder combines global multi-view information, frame-level features, temporal embeddings, physical latents, patch-level features, and multi-scale spatial RGB conditions. These conditions are first projected into a low-resolution seed feature map, then refined by temporal mixing blocks and upsampling layers to produce a shared dense feature tensor:
\begin{equation}
G = D_{\mathrm{shared}}(z, f_{v,t}, z_{\mathrm{phys}}, c_{v,t}, p_{v,t}),
\end{equation}
where \(c_{v,t}\) denotes multi-scale spatial RGB conditions and \(p_{v,t}\) denotes patch-token conditions. The shared tensor \(G\) is then used by task-adaptive heads to predict flow, force, and stress.

Each task head modulates the shared feature using its corresponding physical latent:
\begin{equation}
\hat{Y}_k = H_k(G, h_k), \qquad k \in \{f,m,s\},
\end{equation}
where \(\hat{Y}_f\), \(\hat{Y}_m\), and \(\hat{Y}_s\) denote the predicted flow, force/contact, and stress fields, respectively. This design allows the three dense prediction tasks to share common spatiotemporal features while preserving task-specific behavior.

\subsubsection{Sequential Stress Refinement}

Stress is more indirect and difficult to infer from RGB observations than motion or contact. Therefore, we impose a sequential physical bias in the stress branch. The model first predicts flow and force/contact fields, then feeds them back into the stress branch through a fusion module:
\begin{equation}
G_s = F_s\left([G; \hat{Y}_f; \hat{Y}_m]\right),
\qquad
\hat{Y}_s = H_s(G_s, h_s).
\end{equation}
This design encourages stress prediction to be grounded in visible motion and contact evidence. In addition, spatial enhancement blocks are applied to refine flow and stress features. When object masks are available, they are aligned to the prediction resolution and used to suppress background responses, making the predicted physical fields focus on object regions.

\subsection{Field-conditioned Constitutive and Parameter Prediction}

Lightweight spatiotemporal encoders summarize the predicted fields, which are fused with the physics bottleneck for constitutive-type classification and continuous-parameter regression. Thus, the fields are regressor inputs rather than auxiliary outputs alone. We regress in the transformed target space
\begin{equation}
\tilde{p} = [\log(1+E), \nu, \log(1+\rho), \log(1+\sigma_y)].
\end{equation}
The logarithmic transform handles large dynamic ranges, while Poisson's ratio remains in raw space. During replay, $\hat{c}$ selects the elastic, von Mises, or plasticity-with-softening MPM law configured by $\hat{p}$; an incorrect type is never replaced by ground truth. Yield stress is supervised and evaluated only for yielding ground-truth models and is ignored when the predicted replay model is non-yielding.

Instead of regressing material parameters only from the global visual feature, we further encode the predicted dense physical fields with compact 3D convolutional encoders:
\begin{equation}
u_s = \Phi_s(\hat{Y}_s), \qquad
u_f = \Phi_f(\hat{Y}_f), \qquad
u_m = \Phi_m(\hat{Y}_m).
\end{equation}
The field summaries are fused with the physics bottleneck to obtain a field-aware parameter representation:
\begin{equation}
z_p = \Phi_{\mathrm{fuse}}([z_{\mathrm{phys}}; u_s; u_f; u_m]).
\end{equation}
An MLP classification head predicts one of the three constitutive families---elastic jelly-like behavior, metal-like von Mises plasticity, or plasticine-like plasticity with softening---and a regression head predicts transformed material parameters:
\begin{equation}
\hat{c}=C(z_p), \qquad
\hat{\tilde{p}} = R(z_p),
\qquad
\tilde{p} = [\log(1+E), \nu, \log(1+\rho), \log(1+\sigma_y)].
\end{equation}
Large-range quantities are handled in log space, while Poisson's ratio is constrained to a physically valid range during prediction. During replay, $\hat{c}$ selects the MPM stress law and return mapping, and $\hat{p}$ configures it. Ground-truth types are never substituted for predicted types in reported replay metrics.

\subsection{Training Objective}

The model is trained with a multi-task objective that combines constitutive classification, parameter regression, dense field supervision, and physics-consistency regularization:
\begin{equation}
\mathcal{L}
=
\mathcal{L}_{\mathrm{reg}}
+
\lambda_c \mathcal{L}_{\mathrm{cls}}
+
\lambda_s \mathcal{L}_{\mathrm{stress}}
+
\lambda_f \mathcal{L}_{\mathrm{flow}}
+
\lambda_m \mathcal{L}_{\mathrm{force}}
+
\lambda_p \mathcal{L}_{\mathrm{phys}}.
\end{equation}
Here $\mathcal{L}_{\mathrm{cls}}$ is cross-entropy over the three constitutive families. The regression loss is computed in the transformed parameter space. Invalid or undefined dimensions are masked: yield stress is supervised only for samples whose ground-truth law includes yielding, and purely elastic samples are excluded from yield-stress parameter-error statistics. During replay, yield stress is ignored if the predicted type is non-yielding; if an elastic sample is misclassified as yielding, the predicted yield stress is used and the resulting replay error remains included. For dense fields, stress and flow are supervised with foreground-weighted reconstruction losses and edge-consistency regularization. Force prediction is supervised using a combination of binary cross-entropy and Dice loss.

The physics-consistency term encourages agreement among flow, force, and stress responses. It is defined as a sum of three weak structural constraints over the predicted dense fields:
\begin{equation}
\mathcal{L}_{\mathrm{phys}}
=
\mathcal{L}_{sf}
+
\mathcal{L}_{fs}
+
\mathcal{L}_{ff},
\end{equation}
where \(\mathcal{L}_{sf}\) encourages stress and flow fields to be positively correlated, while \(\mathcal{L}_{fs}\) and \(\mathcal{L}_{ff}\) encourage stress and flow responses to be stronger in force/contact regions than in non-contact regions.

Specifically, stress and flow maps are first converted to magnitude maps and normalized with a per-sample z-score. We then penalize low stress-flow correlation with a softplus margin loss:
\begin{equation}
\mathcal{L}_{sf}
=
\mathrm{softplus}(m_{sf} - \mathrm{corr}(\hat{S}, \hat{F})).
\end{equation}
For force-aware consistency, the force map is converted into a soft contact weight map, and we impose a contrastive margin between the average field magnitude inside and outside the contact region:
\begin{equation}
\mathcal{L}_{fs}
=
\mathrm{softplus}
\left(
m_{fs} -
(\mu_{\mathrm{in}}(\hat{S}) - \mu_{\mathrm{out}}(\hat{S}))
\right),
\end{equation}
\begin{equation}
\mathcal{L}_{ff}
=
\mathrm{softplus}
\left(
m_{ff} -
(\mu_{\mathrm{in}}(\hat{F}) - \mu_{\mathrm{out}}(\hat{F}))
\right).
\end{equation}

\subsubsection{Target-space Transformation and Validity Masking}

We do not regress the raw physical values directly. Instead, we transform the target parameters as
\begin{equation}
\tilde{\mathbf{p}} =
\left[
\log(1+E),\,
\nu,\,
\log(1+\rho),\,
\log(1+\sigma_y)
\right].
\end{equation}
This transformation compresses large dynamic ranges and improves optimization stability. We further construct a per-dimension validity mask so that only physically meaningful targets participate in the loss. Specifically, we require $E>0$, $\rho>0$, and $\nu \in (0,0.5]$; yield stress additionally requires $\sigma_y>0$ and a yielding ground-truth constitutive family. The transformed targets are then normalized using training-set statistics before regression loss is applied.

\subsection{Staged Optimization}

To reduce gradient competition among tasks, we adopt a staged training schedule. The model first learns flow and force prediction, which are more directly observable from RGB videos. It then trains the stress branch conditioned on the learned flow and force fields. After that, the full dense-field pipeline is optimized jointly. Finally, the dense-field components are frozen and the constitutive classification and parameter regression branches are refined using the predicted physical responses.

This training order follows the physical dependency encoded in the architecture: contact and motion provide low-level evidence, stress is inferred from these responses, and the constitutive family and continuous parameters are estimated from the resulting physical state.

\section{Experimental Setup and Evaluation Protocol}
\label{sec:evaluation-protocol}
\subsection{Data Organization}
Each training sample contains RGB, stress, flow, force-mask, and object-mask videos, together with action labels and four-dimensional material parameters. The dataset is organized as aligned multi-view spatiotemporal tensors, ensuring efficient joint training for field prediction and parameter inference.

\subsection{Baselines}
We compare with (1) a \textbf{VLM-based estimator}, motivated by multimodal and zero-shot physical reasoning~\cite{xuGaussianPropertyIntegratingPhysical2024,shuaiPUGSZeroshotPhysical2025,zhaiPhysicalPropertyUnderstanding2024}, that predicts material and motion types plus four parameters from ordered frames; and (2) a \textbf{supervised video baseline} with the same DINO backbone and multi-view fusion but no physics bottleneck or field-conditioned regression. The latter predicts fields and parameters from a shared representation, isolating the contribution of bridging through predicted fields.
\subsection{Metrics Used in Tables}

Material parameters have different units and dynamic ranges. We therefore evaluate $E$, density $\rho$, and yield stress $\sigma_y$ in $\log(1+x)$ space, and evaluate Poisson's ratio $\nu$ in raw space. For each parameter, we report median absolute error (MedAE), which is more robust than mean error under heavy-tailed failures. We also report sample-wise parameter error (SPE) to measure the quality of the whole predicted parameter set for each simulation sample. For parameter $k$, we normalize its absolute error by the central $80\%$ ground-truth range $r_k$, obtained after removing the lowest and highest $10\%$ ground-truth values:
\begin{equation}
\mathrm{SPE}^{(i)}
= \frac{1}{|\mathcal{K}_i|}
\sum_{k\in\mathcal{K}_i}
\frac{|\hat{y}_{i,k}-y_{i,k}|}{r_k},
\end{equation}
where $\mathcal{K}_i$ contains valid parameters for sample $i$. Yield stress is excluded for purely elastic ground-truth samples. Unfiltered metrics use all test samples. Top-20\% filtered metrics remove the 20\% samples with the largest SPE for each method. This filtering is used because the dataset contains metal objects whose stiffness, especially $E$, is visually ambiguous and causes large errors for all methods. The filtered scores describe typical deformable-object inference, while the unfiltered scores remain the primary evidence of failure sensitivity. Because errors for $E$, $\rho$, and $\sigma_y$ are measured after a $\log(1+x)$ transform, a log-space MedAE such as 1.521 still represents substantial absolute uncertainty and should not be read as precise parameter recovery. The replay evaluation complements, rather than replaces, this parameter-space limitation.

For fields, stress is evaluated mainly by hotspot recall because stress localization is more important than exact full-field reconstruction. Let $S$ and $\hat{S}$ be the ground-truth and predicted stress magnitude maps inside the object mask. We define the hotspot region as the top-response region,
\begin{equation}
\Omega_S = \{x \mid S(x) \ge q_{\tau_S}(S)\}, \quad
\hat{\Omega}_S = \{x \mid \hat{S}(x) \ge q_{\tau_S}(\hat{S})\},
\end{equation}
where $q_{\tau_S}$ denotes a high-percentile threshold. Hotspot recall is
\begin{equation}
\mathrm{HotspotRecall}
=
\frac{|\hat{\Omega}_S \cap \Omega_S|}{|\Omega_S|}.
\end{equation}
It measures whether the predicted stress field recovers the true concentrated stress region. Region IoU uses the same hotspot masks to measure overlap, while soft IoU computes overlap on normalized continuous stress responses. Flow is evaluated by moving-region coverage and velocity-direction agreement, summarized by motion quality. Force is a binary contact mask, so we directly evaluate masked reconstruction and region overlap.

\section{Experiments}

We evaluate parameter and constitutive inference, dense fields, simulation replay, and protocol-matched transfer; the evaluation also covers scaling, unseen interactions, domain adaptation, input robustness, and real videos. Unless noted otherwise, all supervised methods use three-view RGB videos and the cleaned object-level split of 5,000 samples: 496 valid samples for $E$, $\nu$, and density, and 164 yielding samples for yield stress.

\subsection{Benchmark Comparison}

Table~\ref{tab:benchmark_main} compares our method with the VLM and supervised baselines on simulation-ready material parameter inference. We report per-parameter MedAE as well as sample-wise parameter error (SPE), which normalizes each sample's valid parameter errors by robust test-set ranges.

\begin{table}[tbp]
\centering
\caption{Comparison with VLM and direct regression baselines on simulation-ready material parameter inference. 
We report unfiltered parameter-level median absolute error (MedAE) and sample-wise parameter error (SPE), together with SPE after removing the top 20\% highest-SPE samples for each method. 
$E$, density, and yield stress are evaluated in log-space, while $\nu$ is evaluated in raw space.}
\scriptsize
\setlength{\tabcolsep}{1mm}
\label{tab:benchmark_main}
\begin{adjustbox}{max width=\linewidth}
\begin{tabular}{lcccccccc}
\toprule
\multirow{2}{*}{Method} & \multicolumn{4}{c}{MedAE} & \multicolumn{4}{c}{SPE}\\
\cmidrule(lr){2-5}
\cmidrule(lr){6-9}
~ 
& $E$ $\downarrow$ 
& $\nu$ $\downarrow$ 
& Density $\downarrow$ 
& Yield $\downarrow$ 
& \shortstack{Unfilt.\\Med. $\downarrow$}
& \shortstack{Unfilt.\\Mean $\downarrow$}
& \shortstack{Top-20\% Filt.\\Med. $\downarrow$}
& \shortstack{Top-20\% Filt.\\Mean $\downarrow$} \\
\midrule
VLM baseline 
& 2.601 & 0.079 & 1.301 & 3.028 & 0.645 & 0.677 & 0.600 & 0.576 \\
Supervised baseline
& 1.901 & 0.048 & 0.408 & 1.797 & 0.304 & \textbf{0.314} & 0.275 & 0.270 \\
Ours
& \textbf{1.521}
& \textbf{0.045}
& \textbf{0.389}
& \textbf{1.585}
& \textbf{0.299}
& 0.334
& \textbf{0.251}
& \textbf{0.266} \\
\bottomrule
\end{tabular}
\end{adjustbox}%
\end{table}
The VLM baseline is weak at numerical recovery, whereas supervised regression is competitive and has the best unfiltered mean SPE. Relative to that baseline, ours reduces MedAE from 1.901 to 1.521 for $E$, 0.048 to 0.045 for $\nu$, 0.408 to 0.389 for density, and 1.797 to 1.585 for yield stress; it also has the best median and top-20\%-filtered SPE. The supervised baseline's better mean SPE (0.314 versus 0.334) exposes a heavier tail for ours, primarily on visually ambiguous high-stiffness metal, rather than uniform superiority. Sections~\ref{sec:parameter-visualization} and~\ref{sec:material-analysis} provide the full distributions and material-wise analysis.

\subsubsection{Parameter Prediction Visualization}
\label{sec:parameter-visualization}
\begin{figure}[tbp]
    \centering
    \includegraphics[width=\textwidth]{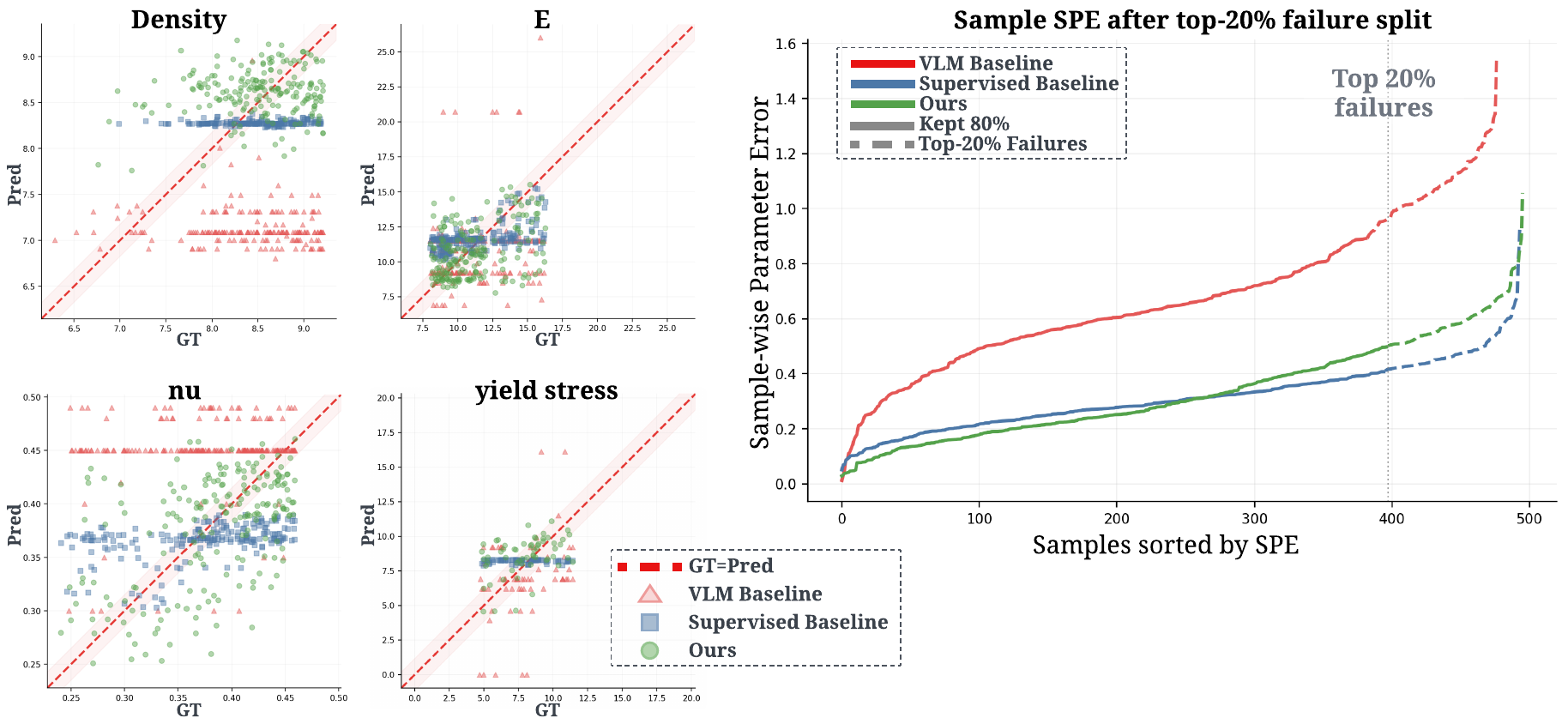}
    \caption{Material-parameter predictions from the VLM baseline, supervised baseline, and our model. The four scatter plots show prediction versus ground truth for density, $E$, $\nu$, and yield stress; the red diagonal is ideal agreement. The sorted SPE plot reports the full sample-wise error distribution, with dashed segments identifying each method's highest-error 20\%. It exposes both the improved typical-sample performance and the remaining long-tail failures.}
    \label{fig:predict_gt}
\end{figure}

Figure~\ref{fig:predict_gt} provides additional visualizations for material-parameter prediction. The scatter plots compare predicted and ground-truth values for each parameter, while the sorted SPE curve shows the distribution of per-sample parameter-set errors and highlights the top-20\% highest-error cases.

\subsubsection{Material-wise Analysis}
\label{sec:material-analysis}
A material-wise breakdown further confirms that the long-tail behavior is mainly caused by high-stiffness metal samples. For our method, metal has a much higher SPE than the other material groups, with median/mean SPE of $0.382$, compared with $0.286$ for jelly and $0.252$ for plasticine. The difference is especially pronounced for Young's modulus: the metal subset has an $E$ MedAE of $3.195$ and mean AE of $3.354$, while jelly and plasticine have much lower $E$ MedAE values of $1.137$ and $1.226$, respectively. This supports our observation that visually estimating stiffness for metal is intrinsically difficult under the available deformation cues. Therefore, the unfiltered mean SPE is strongly affected by a small number of high-stiffness metal failure cases, whereas the top-20\% filtered metrics better reflect performance on the common deformable-object regime.
\begin{table}[tbp]
\centering
\caption{Material-wise parameter error analysis of our method. SPE summarizes the normalized error of the full parameter set for each sample. $E$ error is reported in log-space.}
\label{tab:material_breakdown}
\scriptsize
\setlength{\tabcolsep}{6mm}
\begin{adjustbox}{max width=\linewidth}
\begin{tabular}{lccccc}
\toprule
Material 
& Count 
& Med. SPE $\downarrow$ 
& Mean SPE $\downarrow$ 
& $E$ MedAE $\downarrow$ 
& $E$ Mean AE $\downarrow$ \\
\midrule
Jelly 
& 170 & 0.286 & 0.317 & 1.137 & 1.869 \\
Metal 
& 164 & 0.382 & 0.383 & 3.195 & 3.354 \\
Plasticine 
& 162 & \textbf{0.252} & \textbf{0.300} & 1.226 & \textbf{1.359} \\
\bottomrule
\end{tabular}
\end{adjustbox}
\end{table}

\subsection{Physical-Field Evidence}

Physical-field bridging raises stress-hotspot recall from 0.0762 to 0.5121 and flow coverage IoU from 0.4054 to 0.6675 over direct field regression. Without retraining, masking stress raises $E$ MedAE from 1.521 to 2.431 and yield-stress MedAE from 1.585 to 2.134; flow and contact are complementary. Sections~\ref{sec:field-analysis} and~\ref{sec:qualitative-fields} report the complete intervention and field results.

\subsubsection{Quantitative Field Evaluation}
\label{sec:field-analysis}

Beyond parameter regression, we evaluate stress, flow, and force predictions in Table~\ref{tab:field_eval}. These metrics are designed around the physical meaning of each field. For stress, exact full-field reconstruction is less important than identifying high-response regions, since stress hotspots reveal concentrated force and deformation-sensitive areas. We therefore emphasize hotspot recall and region-overlap metrics. For flow, correct coverage of moving regions is not sufficient: the predicted motion should also carry meaningful velocity direction and magnitude. We therefore report both coverage IoU and velocity-aware motion quality. Force is a contact/force mask, so its evaluation is more direct and uses masked reconstruction and region overlap.

Our model improves stress hotspot recall, stress region overlap, flow coverage, velocity overlap, and temporal consistency over the supervised baseline. Compared with the supervised baseline, which directly predicts all outputs from shared video features, our method performs stress inference through a chained physical reasoning process. The predicted flow and force representations provide additional cues about motion discontinuities, contact regions, and force concentration, all of which are closely related to stress localization. Moreover, our model uses both static geometric features and dynamic deformation features, allowing stress prediction to depend on object structure as well as temporal response. These additional physical and spatiotemporal cues make stress hotspots easier to identify, explaining the gain over direct supervised field regression. Qualitative examples of the predicted flow, force/contact, and stress fields are provided in Section~\ref{sec:qualitative-fields}.

\begin{table}[tbp]
\centering
\caption{Field prediction quality on the evaluation split. Metrics are computed inside the object mask. Our physical-field-bridged model achieves stronger hotspot localization and motion-field quality.}
\label{tab:field_eval}
\scriptsize
\setlength{\tabcolsep}{1mm}
\begin{adjustbox}{max width=\linewidth}
\begin{tabular}{lccc cccc c}
\toprule
\multirow{2}{*}{Method}
& \multicolumn{3}{c}{Stress}
& \multicolumn{4}{c}{Flow}
& \multicolumn{1}{c}{Force} \\
\cmidrule(lr){2-4}
\cmidrule(lr){5-8}
\cmidrule(lr){9-9}
& \shortstack{Hotspot\\Rec.$\uparrow$}
& \shortstack{Region\\IoU$\uparrow$}
& \shortstack{Soft\\IoU$\uparrow$}
& \shortstack{Coverage\\IoU$\uparrow$}
& \shortstack{Velocity\\Overlap$\uparrow$}
& \shortstack{Motion\\Quality$\uparrow$}
& \shortstack{Temporal\\Cons.$\uparrow$}
& MSE$\downarrow$ \\
\midrule
Supervised baseline
& 0.0762
& 0.0245
& 0.4815
& 0.4054
& 0.1232
& 0.2184
& 0.7141
& 0.0601 \\
Ours
& \textbf{0.5121}
& \textbf{0.1524}
& \textbf{0.7719}
& \textbf{0.6675}
& \textbf{0.1725}
& \textbf{0.3316}
& \textbf{0.8135}
& \textbf{0.0597} \\
\bottomrule
\end{tabular}
\end{adjustbox}
\end{table}

\subsubsection{Field Contribution Intervention}

To test whether the predicted fields are actually used by the final regressor, we mask one field at inference time without retraining. When flow or contact is masked, the original predicted stress is retained so that the intervention measures the direct contribution of the masked field rather than its indirect effect through the stress branch. Table~\ref{tab:field_masking} shows complementary roles: stress is the strongest cue for $E$ and yield stress, while flow and contact contribute to all four parameters.

\begin{table}[tbp]
\centering
\caption{Inference-time physical-field masking. Values are MedAE; lower is better.}
\label{tab:field_masking}
\scriptsize
\setlength{\tabcolsep}{3.2mm}
\begin{adjustbox}{max width=\linewidth}
\begin{tabular}{lcccc}
\toprule
Regressor input & $E$ & $\nu$ & Density & Yield \\
\midrule
Full model & \textbf{1.521} & 0.045 & \textbf{0.389} & \textbf{1.585} \\
Stress masked & 2.431 & \textbf{0.043} & 0.394 & 2.134 \\
Flow masked & 2.035 & 0.058 & 0.405 & 1.913 \\
Contact masked & 1.980 & 0.057 & 0.406 & 1.878 \\
All fields masked & 2.525 & 0.045 & 0.391 & 2.149 \\
\bottomrule
\end{tabular}
\end{adjustbox}
\end{table}

\subsubsection{Qualitative Field Prediction Results}
\label{sec:qualitative-fields}

\begin{figure}[tbp]
    \centering
    \includegraphics[width=\textwidth]{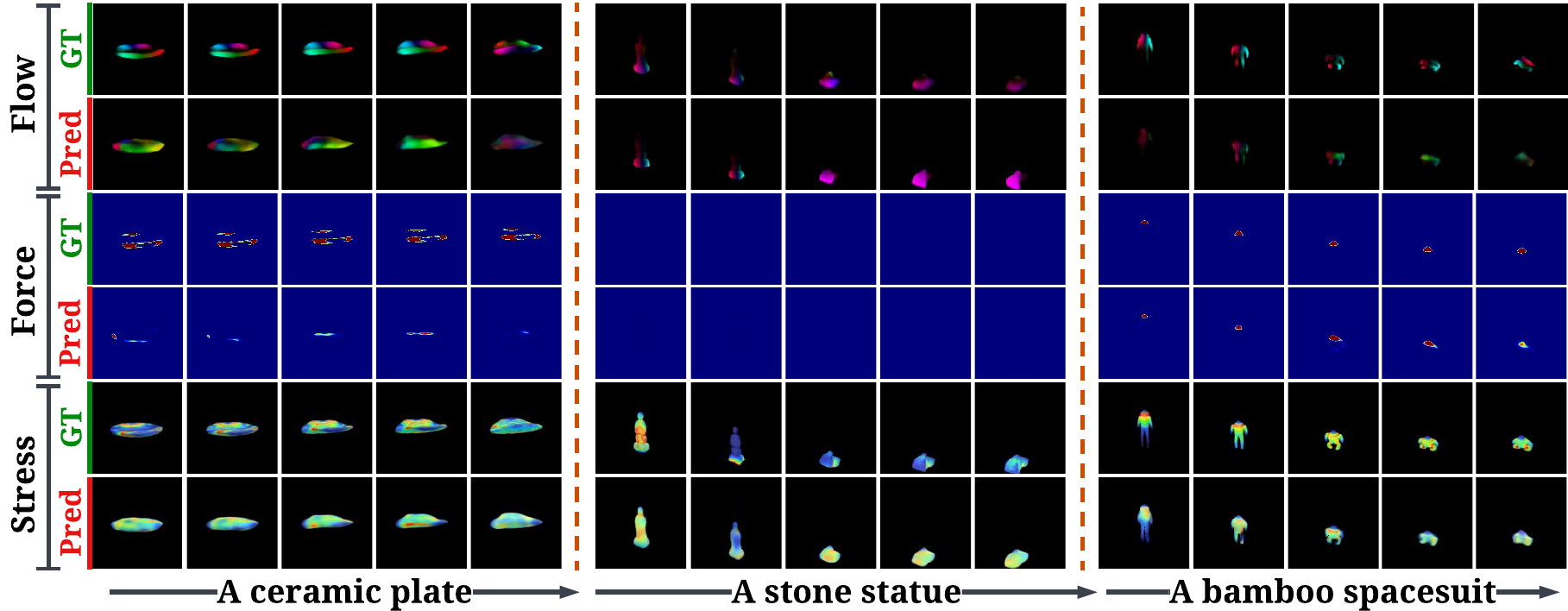}
    \caption{Qualitative physical-field predictions for three unseen objects over representative timesteps. Rows compare ground truth (green label) and prediction (red label) for projected motion flow, force/contact response, and projected stress. The model recovers the principal moving regions, localized contacts, and high-response stress structure despite changes in geometry and action.}
    
    \label{fig:field}
\end{figure}

Figure~\ref{fig:field} shows qualitative examples of the predicted dense physical fields. Across different unseen objects, the model recovers the main motion-flow regions, contact/force regions, and stress-concentration patterns from multi-view RGB videos.

\subsection{Object-level Generalization and Ablation Study}
\label{sec:object-generalization}

We use an object-level train/test split to test whether the model learns transferable physical rules rather than memorizing object-specific appearance. Under this setting, our method achieves the strongest robust sample-wise parameter accuracy. In particular, Table~\ref{tab:ablation} shows that the full model obtains the best unfiltered median SPE and the best top-20\% filtered SPE. Although the ``w/o multi-stage training'' variant has a slightly lower unfiltered Param Mean MAE, it performs worse in SPE, indicating that per-parameter averages alone do not fully capture whether a method predicts a coherent parameter set for each sample.

The ablations confirm the importance of the full design. Removing boundary reasoning weakens edge/contact-sensitive supervision. Removing field auxiliary learning degrades the physical intermediate representation. Removing multi-stage training reduces the stability of the field-bridged learning process. The full model achieves the best range-balanced parameter-set prediction, validating the role of explicit physical fields and staged reasoning in object-level generalization.

\begin{table}[tbp]
\centering
\caption{Ablation study on physical-field-bridged parameter inference. We report unfiltered sample-wise parameter error (SPE) and parameter mean MAE, together with the same metrics after removing the top 20\% highest-SPE samples for each variant. Lower is better.}
\label{tab:ablation}
\scriptsize
\setlength{\tabcolsep}{1.5mm}
\begin{adjustbox}{max width=\linewidth}
\begin{tabular}{lcccccc}
\toprule
\multirow{2}{*}{Variant} 
& \multicolumn{4}{c}{SPE} 
& \multicolumn{2}{c}{MAE} \\
\cmidrule(lr){2-5} \cmidrule(lr){6-7}
& \shortstack{Unfilt.\\Med. $\downarrow$} 
& \shortstack{Unfilt.\\Mean $\downarrow$} 
& Filt. Med. $\downarrow$
& Filt. Mean $\downarrow$
& \shortstack{Unfilt. Param\\Mean $\downarrow$}
& \shortstack{Filt. Param\\Mean $\downarrow$} \\
\midrule
w/o boundary reasoning 
& 0.321 & 0.345 & 0.272 & 0.279 & 1.197 & 1.013 \\
w/o field auxiliary learning 
& 0.312 & 0.349 & 0.266 & 0.274 & 1.204 & 1.023 \\
w/o multi-stage training 
& 0.326 & 0.349 & 0.275 & 0.285 & \textbf{1.118} & 0.946 \\
Ours 
& \textbf{0.299} 
& \textbf{0.334} 
& \textbf{0.251}
& \textbf{0.266}
& 1.139 
& \textbf{0.940} \\
\bottomrule
\end{tabular}
\end{adjustbox}
\end{table}

\subsection{Simulation-based Evaluation under Parameter Ambiguity}

\begin{table}[tbp]
\centering
\caption{Prediction quality under parameter ambiguity. Results are averaged over 10 unseen objects, five loading types, and 5,000 parameter combinations.}
\label{tab:ours_quality}
\scriptsize
\setlength{\tabcolsep}{1mm}
\begin{adjustbox}{max width=\linewidth}
\begin{tabular}{llcccccc}
\toprule
Method & Metric & Bend & Drop & Press & Shear & Stretch & Mean \\
\midrule
Supervised
& SSIM $\uparrow$
& 0.736 & 0.628 & 0.762 & 0.732 & 0.765 & 0.725 \\
& PSNR $\uparrow$
& 24.22 & 22.25 & 24.77 & 23.78 & 24.20 & 23.86 \\
\midrule
Ours
& SSIM $\uparrow$
& 0.918 & 0.746 & 0.845 & 0.921 & 0.907 & 0.868 \\
& PSNR $\uparrow$
& 39.01 & 27.78 & 33.68 & 46.34 & 37.36 & 36.82 \\
\bottomrule
\end{tabular}
\end{adjustbox}
\end{table}
Because distinct parameters can produce similar trajectories under limited actions, we replay the predicted constitutive type and parameters under the observed action and compare renderings. Misclassified types are not corrected. Table~\ref{tab:ours_quality} shows higher replay SSIM and PSNR than the supervised baseline across all five modes, but these scores measure behavior rather than exact parameter recovery; Section~\ref{sec:replay-analysis} provides additional analysis.

\subsubsection{Simulation Replay Analysis}
\label{sec:replay-analysis}

\begin{figure}[tbp]
    \centering
    \includegraphics[width=\linewidth]{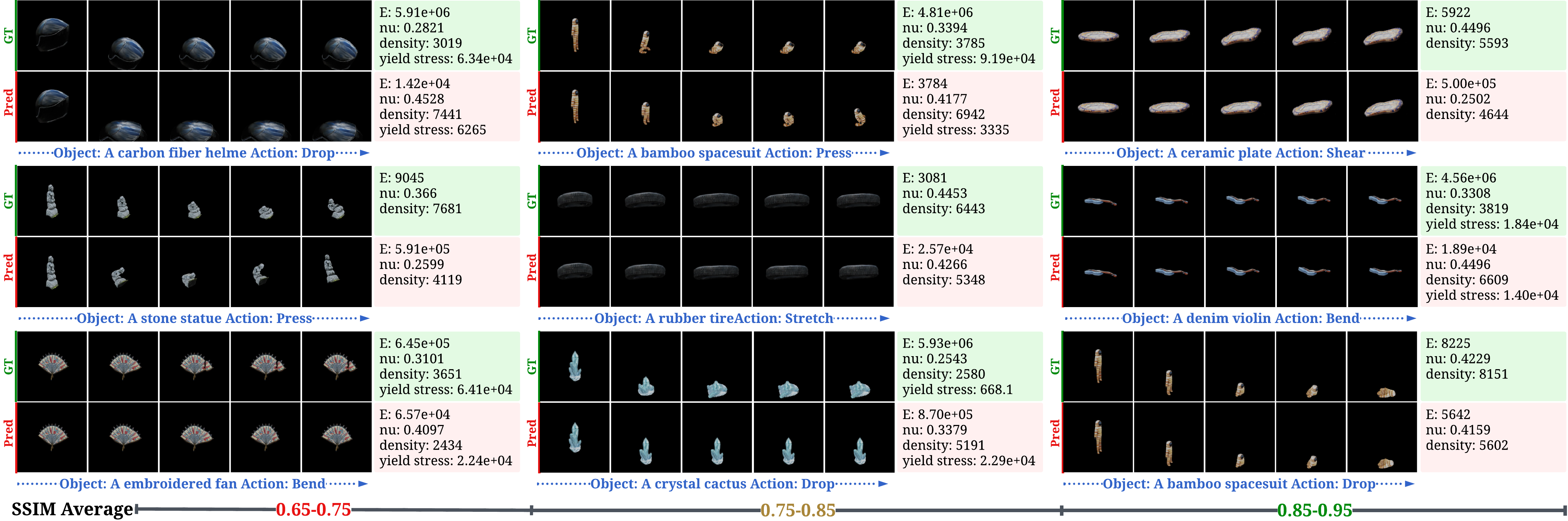}
    \caption{
    Simulation replay examples across unseen objects and loading modes, grouped from lower to higher SSIM. Each example compares ground-truth and predicted trajectories and lists the corresponding continuous parameter sets. Similar motion can arise from noticeably different parameters, illustrating why replay agreement complements but does not replace parameter-space evaluation.
    }
    \label{fig:app_replay_quantitative}
\end{figure}

Figure~\ref{fig:app_replay_quantitative} provides additional quantitative analysis of the simulation-based replay results. This analysis complements Table~\ref{tab:ours_quality} by showing the distribution of replay quality across unseen objects and loading modes.

\subsection{Protocol-Matched Cross-Benchmark Evaluation}

Under Vid2Sim's GSO protocol (16 inference and 8 prediction frames), Vid2Sim reaches 25.07 dB/0.945 PSNR/SSIM and Spring-Gaus reaches 18.32 dB/0.905, while ours reaches 17.15 dB/0.882 in about 30 seconds rather than 15--120 minutes. On Spring-Gaus \cite{zhong2024springgaus}, its native method reaches 17.06 dB/0.897 and ours 12.35 dB/0.791. This is an accuracy--efficiency trade-off, not cross-simulator superiority: Vid2Sim uses roughly 50K matched falling-object sequences, and interaction and simulator shifts reduce transfer. Section~\ref{sec:external-benchmarks} gives the protocol, gap diagnosis, and adaptation.

\label{sec:external-benchmarks}

\subsubsection{GSO Future-State Protocol}

Following Vid2Sim, we use the first 16 frames for material inference, align the simulated state with the final observed frame, and predict the subsequent 8 frames. Table~\ref{tab:gso_protocol} reports all methods on this common protocol. Runtimes are measured or reported on one NVIDIA RTX 4090.

\begin{table}[tbp]
\centering
\caption{Protocol-matched evaluation on GSO.}
\label{tab:gso_protocol}
\scriptsize
\setlength{\tabcolsep}{5mm}
\begin{adjustbox}{max width=\linewidth}
\begin{tabular}{lccc}
\toprule
Method & PSNR $\uparrow$ & SSIM $\uparrow$ & Runtime/sequence $\downarrow$ \\
\midrule
Vid2Sim~\cite{chen2025vid2sim} & 25.07 & 0.945 & $\sim$15--120 min \\
Spring-Gaus~\cite{zhong2024springgaus} & 18.32 & 0.905 & $\sim$15--120 min \\
Ours & 17.15 & 0.882 & $\sim$30 s \\
\bottomrule
\end{tabular}
\end{adjustbox}
\end{table}

Vid2Sim is trained on approximately 50K falling-object sequences closely aligned with the GSO evaluation, whereas our original training set spans multiple actively applied forces and is not specialized to dropping. Per-scene approaches can specialize reconstruction and parameters to each sequence; our method instead amortizes inference across scenes. The result should therefore be interpreted as an accuracy--efficiency trade-off rather than an accuracy claim over per-scene optimization.

\subsubsection{Complete Spring-Gaus Synthetic Set}

We retain every Spring-Gaus sequence rather than selecting simulator-compatible cases. Spring-Gaus obtains 17.06 dB PSNR and 0.897 SSIM, while our zero-shot model obtains 12.35 dB and 0.791. Directly replaying the reported Spring-Gaus parameters in our simulator does not reproduce all benchmark trajectories: several objects rebound above 60\% of their initial drop height, whereas the rebound in our original configuration remains below approximately 5\% even after sweeping the material ranges represented in DeformFieldBench. The discrepancy indicates that trajectories depend on damping, interior filling, time integration, contact implementation, and boundary-condition implementation in addition to nominal material parameters. Complete reproduction of all Spring-Gaus simulator settings is not possible from the available implementation details.

\subsubsection{Lightweight Domain Adaptation}

To separate method capacity from domain mismatch, we generate two domain-specific sets: 500 falling-object simulations following the Vid2Sim generation protocol, and 500 simulations configured to better reproduce Spring-Gaus dynamics. The original model is fine-tuned separately on each set.

\begin{table}[tbp]
\centering
\caption{Effect of 500-simulation domain adaptation.}
\label{tab:domain_adaptation}
\scriptsize
\setlength{\tabcolsep}{4mm}
\begin{adjustbox}{max width=\linewidth}
\begin{tabular}{lcc}
\toprule
Evaluation benchmark & Zero-shot PSNR/SSIM & Adapted PSNR/SSIM \\
\midrule
Vid2Sim/GSO & 17.15 / 0.882 & 20.21 / 0.923 \\
Spring-Gaus & 12.35 / 0.791 & 13.97 / 0.824 \\
\bottomrule
\end{tabular}
\end{adjustbox}
\end{table}

Adaptation improves GSO by 3.06 dB and 0.041 SSIM and Spring-Gaus by 1.62 dB and 0.033 SSIM. These improvements indicate a substantial interaction- and simulator-domain component in the zero-shot gap, but they do not eliminate the gap or prove simulator invariance. Broader randomization over damping, object filling, time integration, contact, and boundary conditions is an important extension; it is not retroactively attributed to the original 5K training results.

\subsubsection{Broader Robustness and Real-world Transfer}

On 50 disjoint DiffSplat objects, ours reaches 36.92 dB/0.8820 PSNR/SSIM; full-set training improves over 10 objects by 13.41 dB/0.1635. Without interaction-specific retraining, squeezing/lifting and force-controlled indentation reach 36.41/0.842 and 24.10/0.712, while five real sequences span 15.864--20.561 dB and 0.7714--0.8430. Sections~\ref{sec:expanded-generalization}--\ref{sec:real-world} include complete results and view/resolution sensitivity. These tests measure behavior-level transfer under unknown real forces, heterogeneous composition, viewpoint shift, and simulator mismatch.

\subsection{Expanded Object and Interaction Generalization}
\label{sec:expanded-generalization}

\subsubsection{Additional Unseen DiffSplat Objects}

We generate 50 objects with DiffSplat~\cite{linDiffSplatRepurposingImage2025} that are disjoint from the original 100 assets. All objects, including severe failures, are retained in the aggregate results.

\begin{table}[tbp]
\centering
\caption{Continuous-parameter accuracy on 50 additional unseen objects. Values are MedAE.}
\label{tab:diffsplat_parameters}
\scriptsize
\setlength{\tabcolsep}{5mm}
\begin{adjustbox}{max width=\linewidth}
\begin{tabular}{cccc}
\toprule
$E$ $\downarrow$ & $\nu$ $\downarrow$ & Density $\downarrow$ & Yield $\downarrow$ \\
\midrule
1.620 & 0.039 & 0.374 & 1.821 \\
\bottomrule
\end{tabular}
\end{adjustbox}
\end{table}

The corresponding future-state replay reaches 36.92 dB PSNR and 0.8820 SSIM. Approximately 10\% of the new assets exhibit severe replay failure or geometric collapse, and these cases are not removed. Post-hoc inspection suggests that some failures resemble training geometries whose simulations occupy narrow parameter ranges. This is a diagnostic hypothesis rather than a confirmed causal explanation and may indicate residual geometry--parameter correlation.

\subsubsection{Dataset Scaling}

We train the same model with simulations from only 10 objects and compare it with the full training set.

\begin{table}[tbp]
\centering
\caption{Replay performance as object diversity increases.}
\label{tab:dataset_scaling}
\scriptsize
\setlength{\tabcolsep}{7mm}
\begin{adjustbox}{max width=\linewidth}
\begin{tabular}{lcc}
\toprule
Training configuration & PSNR $\uparrow$ & SSIM $\uparrow$ \\
\midrule
10 training objects & 23.41 & 0.7041 \\
Full training set & 36.82 & 0.8676 \\
Improvement & +13.41 & +0.1635 \\
\bottomrule
\end{tabular}
\end{adjustbox}
\end{table}

The large gain confirms that the model benefits from increased diversity in object geometry and simulated trajectories.

\subsubsection{Held-Out Interactions}

We evaluate two interactions absent from the five original training modes, without interaction-specific retraining.

\begin{table}[tbp]
\centering
\caption{Generalization to held-out interactions.}
\label{tab:heldout_interactions}
\scriptsize
\setlength{\tabcolsep}{7mm}
\begin{adjustbox}{max width=\linewidth}
\begin{tabular}{lcc}
\toprule
Interaction & PSNR $\uparrow$ & SSIM $\uparrow$ \\
\midrule
Gripper squeezing and lifting & 36.41 & 0.842 \\
Force-controlled indentation & 24.10 & 0.712 \\
\bottomrule
\end{tabular}
\end{adjustbox}
\end{table}

Squeezing and lifting transfer well because their contact and deformation patterns overlap with structures represented during training. Force-controlled indentation is more distinct and yields lower, but still meaningful, replay agreement. Transfer therefore depends on how well the new interaction is covered by the physical patterns in the training distribution.

\subsection{Input Resolution and View Robustness}
\label{sec:input-robustness}

\subsubsection{Resolution Sensitivity}

We downsample the RGB videos at inference time without retraining or fine-tuning.

\begin{table}[tbp]
\centering
\caption{Resolution sensitivity of continuous-parameter prediction. Values are MedAE.}
\label{tab:resolution_sensitivity}
\scriptsize
\setlength{\tabcolsep}{4mm}
\begin{adjustbox}{max width=\linewidth}
\begin{tabular}{lcccc}
\toprule
Resolution & $E$ $\downarrow$ & $\nu$ $\downarrow$ & Density $\downarrow$ & Yield $\downarrow$ \\
\midrule
$512\times512$ & 1.5210 & 0.0450 & 0.3890 & 1.5850 \\
$256\times256$ & 1.8365 & 0.0638 & 0.4059 & 1.7163 \\
$128\times128$ & 1.9200 & 0.0651 & 0.5083 & 1.6780 \\
\bottomrule
\end{tabular}
\end{adjustbox}
\end{table}

Full-resolution input is strongest overall, but the model remains functional at $128\times128$ without resolution-specific training.

\subsubsection{Number and Configuration of Views}

The original cameras are stationary within each sequence and share three canonical horizontal extrinsics across normalized scenes. Because the architecture expects three inputs, single-view inference repeats one available view three times; two-view inference duplicates one of the two views. The model does not receive camera extrinsics and does not impose camera-transformation equivariance.

\begin{table}[tbp]
\centering
\caption{Single-view inference without retraining. Values are MedAE.}
\label{tab:single_view}
\scriptsize
\setlength{\tabcolsep}{3.5mm}
\begin{adjustbox}{max width=\linewidth}
\begin{tabular}{lcccc}
\toprule
Input configuration & $E$ & $\nu$ & Density & Yield \\
\midrule
Three canonical views & 1.521 & 0.0450 & 0.389 & 1.585 \\
First view repeated & 2.026 & 0.0574 & 0.420 & 1.921 \\
Second view repeated & 1.670 & 0.0494 & 0.357 & 2.047 \\
Third view repeated & 1.580 & 0.0482 & 0.391 & 1.745 \\
\bottomrule
\end{tabular}
\end{adjustbox}
\end{table}

With duplicated two-view inputs, $E$ MedAE ranges from 1.833 to 2.144 depending on the pair and duplicated view. Across random three-view configurations, $E$ MedAE ranges from 1.483 to 2.491 and replay PSNR ranges from 21.67 to 35.42 dB. Several configurations match or slightly improve on the canonical views, but top-down or bottom-up views can hide the dominant contact and deformation. These results show that observable evidence matters at least as much as nominal view count, while the performance range confirms meaningful camera sensitivity.

Future work should randomize camera angle, elevation, view count, view dropping, and ordering during training. A pose-conditioned set encoder that consumes camera extrinsics would support variable view counts more directly; these extensions are not part of the reported original model.

\subsection{Real-world Object Application}
\label{sec:real-world}
We quantitatively evaluate five real deformation sequences involving two previously unseen objects: a wallet under bending, compression, and stretching, and a thin-walled plastic bottle under bending and compression. No real-data fine-tuning or ground-truth force, stress, material parameter, or constitutive-type label is used.

\paragraph{Replay protocol}
For each interaction, we capture synchronized multi-view RGB videos, predict the constitutive type and continuous parameters, align the initial object geometry and cameras, and replay the observed interaction with the predicted material configuration. The same foreground mask is applied to the aligned real and rendered sequences. Masks are initialized with SAM and manually refined to remove background regions and areas occluded by the manipulating hand. We then compute image-space PSNR and SSIM between the real and rendered foreground sequences.

\begin{table}[tbp]
\centering
\caption{Quantitative replay on real-world deformation videos.}
\label{tab:real_world_replay}
\scriptsize
\setlength{\tabcolsep}{7mm}
\begin{adjustbox}{max width=\linewidth}
\begin{tabular}{l l cc}
\toprule
Object & Interaction & PSNR $\uparrow$ & SSIM $\uparrow$ \\
\midrule
Wallet & Bending & 15.864 & 0.7714 \\
Wallet & Compression & 16.427 & 0.7936 \\
Wallet & Stretching & 16.042 & 0.7851 \\
Plastic bottle & Bending & 17.642 & 0.8231 \\
Plastic bottle & Compression & 20.561 & 0.8430 \\
\bottomrule
\end{tabular}
\end{adjustbox}
\end{table}

The plastic-bottle compression sequence has the strongest agreement, where interaction and boundary conditions can be aligned most reliably. Residual differences arise from geometry and camera alignment, unknown contact forces, approximate boundary conditions, hand occlusion, trajectory mismatch, heterogeneous object composition, and constitutive behavior outside the current simulator. PSNR and SSIM therefore measure replay consistency at the deformation-behavior level; they do not establish that the inferred numerical parameters equal unknown real material properties.

\section{Discussion and Limitations}
\label{sec:discussion}

Our results suggest that simulation-ready material inference benefits from explicitly decomposing visual observations into dense physical fields before predicting a constitutive family and global material parameters. Compared with direct video-to-parameter regression, force/contact, motion flow, and stress response provide more interpretable physical evidence, especially when rebound, oscillation, recovery, and stress concentration are subtle.

Despite these benefits, our study has several limitations. First, DeformFieldBench is primarily synthetic. The real-world replay study provides quantitative behavior-level evidence but lacks calibrated force sensing and ground-truth material parameters, so it cannot validate absolute recovery. Second, zero-shot performance drops across simulator and interaction domains. Damping, object filling, integration, contact, and boundary-condition implementations can change a trajectory even when nominal material parameters match. Third, the original model is trained with a canonical three-camera arrangement, does not consume extrinsics, and remains sensitive to whether a view exposes contact and deformation. Fourth, its flow and stress outputs are image-space projections rather than volumetric fields. Finally, continuous-parameter errors remain substantial even when replay appears plausible, reflecting both observation ambiguity and the many-to-one relationship between parameters and trajectories.

\paragraph{Homogeneous-material assumption}
The current output is one global constitutive type and parameter set per object. Dense physical fields act as spatial evidence for global system identification; they do not recover regional material properties. For a composite object such as a wallet made of leather, fabric, stitching, and metal, the result should be interpreted as an effective global approximation of the dominant observed deformation, not as the physical property of each component. A spatial extension could lift multi-view and field features to surface points, volumetric elements, or MPM particles and predict region-level constitutive types and parameters. Such an extension requires heterogeneous training objects, material-boundary annotations, image-to-particle correspondence, region-level supervision, and simulators that handle material interfaces.

We observe that most long-tail parameter errors are concentrated on high-stiffness metallic samples. This failure mode is partly caused by an identifiability limitation of the current loading protocol, rather than only by insufficient model capacity. In our simulator, press, stretch, shear, and bend actions are imposed by kinematic actuators with prescribed trajectories. For high-stiffness metallic objects, this can resemble applying an effectively large external force until deformation occurs. Under such force-saturated loading, metallic objects with different Young's moduli or yield stresses may exhibit similar visible behavior, including limited elastic recovery, permanent deformation, and saturated relative stress responses. Consequently, their RGB deformation trajectories, projected flow fields, and normalized image-space stress maps become less discriminative than those of more recoverable materials such as jelly-like or plasticine-like objects.

A direct way to reduce this ambiguity is to include absolute force or stress measurements. For example, a robotic force-torque sensor, tactile array, pressure sensor, or calibrated actuator-force signal could reveal the force magnitude at which deformation starts, which would be informative for distinguishing near-rigid metallic materials. Our current dataset does not include such absolute external-force measurements or per-particle absolute stress targets; instead, it provides rendered and normalized physical-response fields. This design keeps the benchmark lightweight, visually aligned, and suitable for dense pixel-level supervision, but it also limits the identifiability of high-stiffness materials whose visible deformation responses are weak or saturated.

Future versions of DeformFieldBench could incorporate calibrated loading magnitudes, per-point absolute stress, tactile or force-torque measurements, spatially varying materials, and broader randomization of simulator settings. Training with randomized camera angle, elevation, view count, view dropping, and view order would reduce reliance on the canonical arrangement; a pose-conditioned set encoder could explicitly support arbitrary cameras and variable view counts. These are future extensions rather than components of the original 5K-result model. The framework could also be integrated with vision-language priors and robotic sensing to support downstream reasoning, planning, and interaction.

\section{Conclusion}

We introduced physical-field-bridged material inference, which organizes observable contact, motion, and stress evidence before predicting a constitutive type and continuous parameters from multi-view RGB videos. DeformFieldBench supplies the aligned dense fields and simulator-ready supervision required to study this formulation. The physical fields improve median parameter accuracy, field quality, object-level generalization, and same-simulator replay; protocol-matched, held-out-interaction, input-robustness, adaptation, and real-world evaluations characterize transfer beyond the original split. The current scope remains bounded: absolute parameter recovery is difficult, image-space replay measures behavioral consistency rather than material truth, and performance is sensitive to camera evidence and simulator configuration. Within this scope, the results support physical-field bridging as a practical basis for fast material inference while motivating pose-aware aggregation, broader simulator randomization, calibrated sensing, and spatially varying material prediction.

\FloatBarrier
\bibliographystyle{unsrtnat}
\bibliography{references,references_iclr_additions}
\end{document}